\documentclass[twocolumn,10pt]{asme2ej}

\usepackage{graphicx} %% for loading jpg figures
\usepackage{hyperref}   % to set up hyperlinks
\hypersetup{
	colorlinks=true,
	linkcolor=blue,
	citecolor=blue,
	urlcolor=blue,
}

\usepackage[square,numbers,sort&compress]{natbib}

\usepackage{graphicx}%
\usepackage{subfigure}
\usepackage{float}
\usepackage{amsmath}

\usepackage{xcolor}%
\definecolor{r}{rgb}{0.8,0.1,0.1}
\definecolor{b}{rgb}{0.1,0.1,0.9}
\definecolor{g}{rgb}{0.0,0.3,0.0}

\providecommand{\Rey}{\mathrm{Re}}

\newcommand{\RNum}[1]{\uppercase\expandafter{\romannumeral #1\relax}}

\title{A reduced-scale autonomous morphing vehicle prototype with enhanced aerodynamic efficiency}

\author{Peng Zhang\thanks{Address all correspondence to this author.}\\
    \affiliation{ 
    Department of Mechanical Engineering\\
        Tennessee Technological University\\
        115 W. 10th Street\\
        Cookeville, Tennessee, 38505\\
        Email: pzhang@tntech.edu
    } 
}

\author{Branson Blaylock\\
    \affiliation{ 
    Department of Mechanical Engineering\\
        Tennessee Technological University\\
        115 W. 10th Street\\
        Cookeville, Tennessee, 38505\\
    } 
}

\begin{document}

\maketitle    

%\pagestyle{plain}
%\thispagestyle{plain}
%%%%%%%%%%%%%%%%%%%%%%%%%%%%%%%%%%%%%%%%%%%%%%%%%%%%%%%%%%%%%%%%%%%%%%
\begin{abstract}
{\it
Road vehicles contribute to significant levels of greenhouse gas (GHG) emissions. 
A potential strategy for improving their aerodynamic efficiency and reducing emissions is through active adaptation of their exterior shapes to the aerodynamic environment. 
In this study, we present a reduced-scale morphing vehicle prototype capable of actively interacting with the aerodynamic environment to enhance fuel economy. 
Morphing is accomplished by retrofitting a deformable structure actively actuated by built-in motors. 
The morphing vehicle prototype is integrated with an optimization algorithm that can autonomously identify the structural shape that minimizes aerodynamic drag.  
The performance of the morphing vehicle prototype is investigated through an extensive experimental campaign in a large-scale wind tunnel facility. 
The autonomous optimization algorithm identifies an optimal morphing shape that can elicit an 8.5\% reduction in the mean drag force. 
Our experiments provide a comprehensive dataset that validates the efficiency of shape morphing, demonstrating a clear and consistent decrease in the drag force as the vehicle transitions from a suboptimal to the optimal shape. 
Insights gained from experiments on scaled-down models provide valuable guidelines for the design of full-size morphing vehicles, which could lead to appreciable energy savings and reductions in GHG emissions. 
This study highlights the feasibility and benefits of real-time shape morphing under conditions representative of realistic road environments, paving the way for the realization of full-scale morphing vehicles with enhanced aerodynamic efficiency and reduced GHG emissions.
}
\end{abstract}

%\keywords{Aerodynamics, Autonomous optimization, Genetic algorithm, Morphing structure, Wind tunnel testing}
%%%%%%%%%%%%%%%%%%%%%%%%%%%%%%%%%%%%%%%%%%%%%%%%%%%%%%%%%%%%%%%%%%%%%%
%\begin{nomenclature}
%\entry{A}{Frontal area of the vehicle model}
%\entry{$C_{\mathrm{D}}$}{Drag coefficient}
%\entry{$F_D$}{Drag force}
%\entry{$\overline{F_{\mathrm{D}}}$}{Mean drag force over the acquisition period}
%\entry{$\Rey$}{Reynolds number}
%\entry{$\theta_1$, $\theta_2$, and $\theta_3$}{Local turning angle of the panels}
%\entry{$\Theta$}{Domain of feasible design variables}
%\entry{$\rho$}{Density of air}
%\entry{$U$}{Upstream wind speed}
%\end{nomenclature}

%%%%%%%%%%%%%%%%%%%%%%%%%%%%%%%%%%%%%%%%%%%%%%%%%%%%%%%%%%%%%%%%%%%%%%
\section{Introduction}\label{sec:intro}

In 2022, the transportation section accounted for 28\% of greenhouse gas emissions in the United States, among which 90\% originated from light-duty vehicles and medium- and heavy-duty trucks \citep{EPA}. 
Notably, medium- and heavy-duty vehicles, despite comprising only 10\% of all vehicle miles traveled, contributed to 23\% of transportation-related greenhouse gas (GHG) emissions \citep{EPA}. 
The high emission levels and limited fuel economy of medium- and heavy-duty vehicles could be attributed to their suboptimal aerodynamic designs. 
Their bulky exterior shapes often induce flow separation, creating large, low-pressure wake regions that substantially increase aerodynamic drag
Improving the aerodynamic performance of road vehicles is therefore crucial for enhancing fuel economy and reducing emissions.

Nature provides valuable insights into structurally efficient aerodynamic designs.
Birds, for instance, can dynamically adjust their wings' span, sweep, and camber during flight, adapting to varying speeds and maneuvers to optimize aerodynamic efficiency \citep{Cheney21}. 
During cruising flights, large soaring birds spread their primary feathers to create wingtip slots, which reduce drag by redistributing wingtip vortices \cite{Tucker95}. 
Additionally, many bird species deploy the alula, a small thumb-like feather structure at the wing's leading edge, during slow flight to delay stall by guiding vortices over the wing \cite{Linehan20}. 
Similarly, small covert feathers on the upper wing surface automatically lift at high angles of attack or in gusty conditions, functioning as passive flaps that suppress flow separations \cite{Dvorak16}.

Engineers have drawn inspirations from these biological systems to develop adaptive morphing structures with improved aerodynamic efficiency. 
Aircraft equipped with morphing wings could adjust their camber and curvature during cruise flight, leading to 25\% reductions in drag \cite{Dong25}. 
Likewise, passive deployable flaps inspired by covert feathers have been installed on aircraft wings, which could enhance post-stall control \cite{Sedky24} and reduce unsteady lift forces in gusty airflows \cite{Murayama21}. 
Similar bio-inspired designs have been implemented on road vehicles. 
Surface modifications through flow regulation devices such as flaps, bumps, and vortex generators \citep{Ha10, Bayindirli23, Garcia23, Renani24} have shown notable improvements in vehicles' fuel economy. 
Recently, adaptive aerodynamic structures have been integrated into several commercial vehicle models, such as the retractable spoilers on the Porsche Panamera  \cite{Porsche} and the automated ``centripetal wing'' on the Zenvo TSR-S \cite{Zenvo}, which are capable of adjusting their shapes based on the driving speed.

Despite these promising advancements, flow regulation devices have only led to moderate improvements in road vehicles' aerodynamic efficiency. 
Their limited effectiveness primarily arises from constrained interactions with the airflow due to their small size and lack of extensive morphing capabilities. 
Additionally, their aerodynamic efficiency is further restricted by a lack of dynamic control that enables active adaptation to transient aerodynamic conditions. 
Over the past decades, the concept of morphing vehicles that can actively adapt their exterior shapes to changing aerodynamic environments has garnered increasing attention as a potential avenue for improving fuel economy \citep{Daynes13}. 
Nevertheless, advancements in morphing technologies in the laboratory have not translated into breakthroughs in the automotive industry.
Although futuristic morphing vehicle concepts have emerged in recent years \citep{BMW}, they are still not primed for large-scale production due to the high manufacturing costs associated with their complex and intrusive designs. 

%Over the past few decades, the concept of morphing vehicles that can actively adapt their exterior shapes to the aerodynamic environment has garnered increasing attention as a potential venue for improving fuel economy \citep{Daynes13}. 
%Previous studies have highlighted the significant potential of shape morphing to enhance vehicles' fuel economy \citep{Ando10, Good21}. 
%For example,  \cite{Ando10} employed computational fluid dynamics (CFD) simulations to optimize the shape of a passenger vehicle by extensively altering its exterior geometry, achieving a maximum drag reduction of 16\%. 
%However, advancements in morphing designs within numerical settings have not translated into breakthroughs in the automotive industry.
%Despite the emergence of futuristic morphing vehicle concepts in recent years \citep{BMW}, they are still not primed for large-scale production due to the high costs associated with their complex and intrusive designs. 

In the present work, we put forward an alternative, low-cost design approach to achieve shape morphing. 
Under this design framework, morphing is accomplished by retrofitting the vehicle with an actively controlled deformable structure. 
The feasibility of this design concept has been validated in a computational environment using a combined computational fluid dynamics (CFD) and parametric genetic algorithm framework  \citep{Kazemipour24}, where we identified optimal structural shapes that resulted drag reductions between 8\% -- 10\% across a range of driving speeds. 
This improvement in the vehicle's fuel economy was attributed to the suppression of the flow circulation bubble in the wake, thereby increasing the back-pressure and reducing the aerodynamic  drag. 
%While CFD-based studies have demonstrated the potential of shape morphing, vehicle prototypes with morphing capabilities have yet to be fabricated and their performance remains untested in realistic experimental settings. 
The primary objective of the present study is to demonstrate and validate, for the first time, a morphing vehicle prototype capable of achieving a wide range of exterior shapes in an experimental setting. 
Similar to \cite{Kazemipour24}, our prototype is based on a generic reduced-scale pick-up truck model, due to its significant potential to enhance aerodynamic efficiency through exterior shape optimization.
Morphing is accomplished by retrofitting a simple and cost-effective deformable structure composed of 3D-printed panels actuated by servo motors. 
The prototype is guided by an autonomous optimization and adaptation algorithm to determine the optimal morphing strategy.

Effective control algorithms are essential for enabling the continuous learning and adaptation of morphing vehicles to maximize their energetic benefits. 
While machine learning (ML) algorithms are a popular choice for real-time learning and adaptation \citep{Ramu22}, they often require extensive datasets and cannot guarantee global optimization \cite{Gazzola16}. 
An alternative optimization technique that ensures global optimization is the genetic algorithm (GA) \citep{Goldberg89}, inspired by  the concept of biological evolution. 
In the GA framework,  feasible designs are represented by vector of design parameters \citep{Holland92}. 
A selection process identifies the ``elite'' designs based on their fitness scores, which are then used to produce the next generation of offspring  through genetic operations such as crossover and mutation  \citep{Spears91, Umbarkar15}. 
This iterative process continues until global optimization is achieved.  
GA has been successfully applied to a variety of engineering problems, ranging from optimizing aircraft shapes to designing multi-layered  metamaterials \citep{Lee08, Hilbert06, Madsen00, Saario06, Byrne14}. %Liu17, , Kord24, Cao24
Contrary to ML-based approaches, GA-based control algorithms can quickly evaluate a vehicle's aerodynamic performance and reliably identify a global optimum. 
Our design leverages the GA's rapid convergence and high accuracy to guide the shape deformation of the proposed morphing vehicle prototype.

We embark on an extensive experimental campaign to investigate the performance of the proposed morphing vehicle prototype and the optimization algorithm. 
The experiments are carried out in a large-scale, low-speed wind tunnel (LSWT) facility, capable of generating wind speeds up to $15\,\mathrm{m/s}$, thereby creating an aerodynamic environment that closely resembles realistic driving conditions. 
The LSWT is equipped with a high-frequency load measurement system that can provide precise drag force measurements with fine resolution. 
This state-of-the-art facility ensures accurate and reliable validation of the morphing vehicle's aerodynamic performance and optimization capabilities.

A series of experiments are performed to demonstrate and validate the proposed morphing vehicle concept. 
First, the morphing structure is positioned in the ``neutral'' configuration, where all panels are oriented horizontally. 
This preliminary measurement aims to elucidate the impact of the morphing structure on the vehicle's aerodynamic performance.
Subsequent experiments are designed to evaluate the efficiency of the proposed control algorithm using a ``hardware-in-the-loop'' approach \cite{Boria09}, where wind tunnel testing provides real-time evaluation of the morphing shapes' fitness and the GA-based optimization framework iteratively search for the elite morphing strategy. 
Additional measurements are carried out to validate the accuracy of these elite shapes, whereby we analyze the dynamic variations in the drag force as the morphing structure transitioned from the neutral configuration to the optimal shapes. 
These experiments provide a comprehensive dataset to assess the performance and feasibility of the proposed morphing vehicle concept.

The key novelties of the present work include: 
i)  the development of the first reduced-scale vehicle prototype equipped with a non-invasive morphing design capable of generating a broad range of exterior shapes;
ii) the integration of an efficient optimization algorithm that autonomously learns and identifies the vehicle's shape with optimal aerodynamic efficiency; and 
iii) the acquisition of a comprehensive experimental dataset in the wind tunnel for rigorous validation of the morphing vehicle prototype's performance. 
This study represents a critical first step toward the realization of full-scale morphing vehicles with improved aerodynamic efficiency and reduced GHG emissions, with significant potential for immediate impacts on the automotive industry.

The rest of the paper is organized as follows. 
Section~\ref{sec:method} describes the design of the morphing vehicle prototype, the autonomous optimization algorithm, and the experimental procedures.  
Key findings are presented and discussed in Sec.~\ref{sec:results}. 
Major conclusions and suggestions for future research are summarized in Sec.~\ref{sec:conclude}.

\section{Methods}\label{sec:method}

\subsection{Morphing vehicle prototype}\label{sec:method:fabrication}

\begin{figure}[t]
\centering
\includegraphics[width=0.47\textwidth]{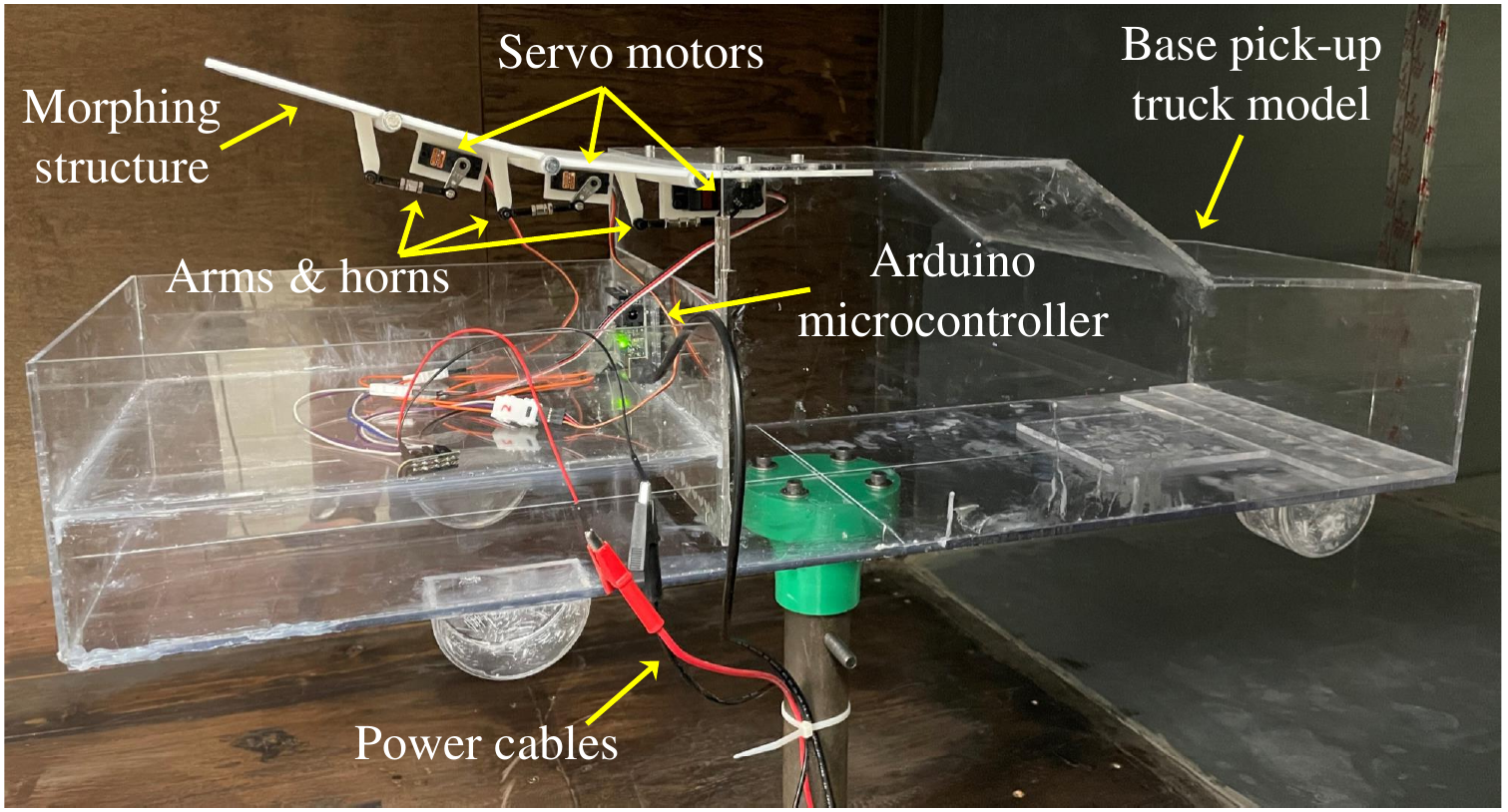}
\caption{Laboratory-scale prototype of the autonomous morphing vehicle
}\label{fig:prototype}
\end{figure}

We embarked on an extensive experimental campaign to demonstrate the proposed morphing vehicle concept and validate the efficiency of the autonomous optimization algorithm. 
To this aim, we fabricated a reduced-scale morphing vehicle model  \citep{Yang05} based on a generic pick-up truck, scaled down at 1:8 ratio. 
The pick-up truck model was selected as the base design due to its significant potential for aerodynamic improvement through body shape optimization.
The length, width, and height of the model measured $78.11\,\mathrm{cm}$,  $25.40\,\mathrm{cm}$, and $20.95\,\mathrm{cm}$, respectively.

The base geometry of the vehicle was constructed using $0.32\,\mathrm{cm}$-thick clear acrylic panels, as shown in Fig.~\ref{fig:prototype}. 
The panels were laser-cut and assembled to form the vehicle's outer surface. 
Silicone glue was applied to seal gaps and prevent airflow leakage through the structure. 
To ensure balanced weight distribution, small acrylic pieces were fixed inside the model, such that the center of mass was positioned at the middle-length of the geometry.

\begin{figure*}[ht]
\centering
\includegraphics[width=0.9\textwidth]{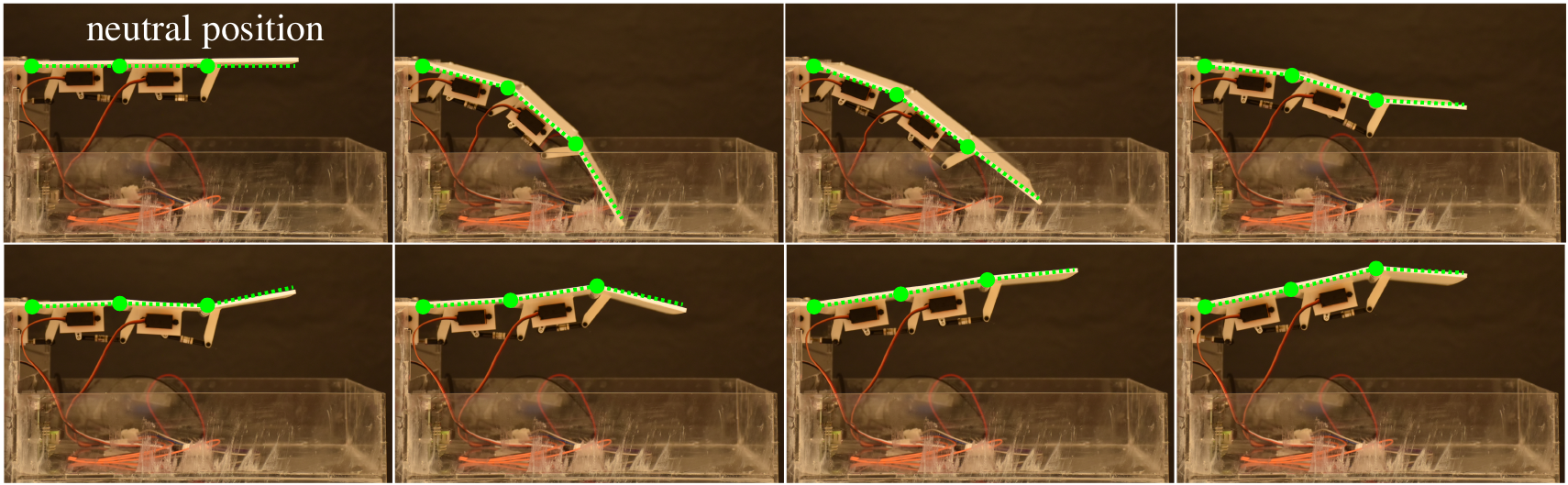}
\caption{A series of exemplary shapes attained by the morphing structure. 
The overlaid green dotted lines illustrate the target morphing shapes. 
The top left image displays the ``neutral'' configuration of the morphing structure, where all panels are oriented horizontally. 
}\label{fig:validate_shapes}
\end{figure*}

Morphing was achieved by installing a deformable structure to the roof of the pick-up truck model. 
The morphing structure consisted of three  rigid plastic panels, each measuring $7.0 \times 24.1\,\mathrm{cm}$ (length$\times$width), which were fabricated using polylactic acid (PLA; eSUN PLA+) on a 3D printer (Bambu Lab; model P1S). 
The width of the panels was intentionally chosen to be slightly smaller than the car's width to avoid occlusion with the bed rails. 
These morphing panels were mounted on a base plate fixed to the roof of the vehicle model. 
Neighboring panels were connected via metal hinges, allowing freely rotations with minimal friction.

Servo motors were installed between panels to control their relative rotation. 
A high-torque servo (Hexfly HX-3225 25KG digital servo; model RER11856) was employed to connect the morphing structure to the base plate, ensuring that it could withstand the weight of the entire morphing structure.  
Light-weight micro servos (Savox micro digital servo; model SH0255MGP) were installed between the morphing panels to provide sufficient torque while minimizing overall structural weight. 
Servo horns and steering rods were employed to adjust the span of the relative rotation angle.  
During actuation, the servo motors were powered by an external power supply providing $5.5\,\mathrm{V}$ direct current.

The  servo motors was actuated by an Arduino micro-controller (Arduino Uno; model Rev3). 
The Arduino board was programmed in MATLAB   (The MathWorks, Inc.; version R2024a)  using its ``Support Package for Arduino Hardware''. 
Prior to our experiments, the servo motors were calibrated to ensure precise control of the morphing panels, correlating the motor rotation angles with the corresponding panel movements.

A series of preliminary tests were performed to validate the accuracy of the morphing control system. 
As shown in Fig.~\ref{fig:validate_shapes}, the morphing structure successfully achieved a wide range of shapes that precisely matched the target configurations specified via MATLAB commands.
To aid our discussions,  the morphing panels were labeled from the fixed to the free end as \RNum{1}, \RNum{2}, and \RNum{3}, and  their relative rotation angles were denoted as $\theta_1$, $\theta_2$, and $\theta_3$, respectively, as illustrated in Fig.~\ref{fig:GA}. 
The admissible rotation angles of the panels are $-20^{\circ} \leq \theta_1 \leq 13^{\circ}$, $-30^{\circ} \leq \theta_2 \leq 15^{\circ}$, and $-30^{\circ} \leq \theta_3 \leq 15^{\circ}$. 
Additionally, the configuration in which all panels were oriented horizontally was defined as the ``neutral'' configuration, as illustrated by the top-left image in Fig.~\ref{fig:validate_shapes}. 
A video showing the dynamic morphing of the proposed vehicle prototype is available in Supplementary Video.

\subsection{Autonomous optimization algorithm}\label{sec:method:algorithm}

The proposed morphing vehicle design concept was integrated with a GA-based autonomous optimization framework. 
Here, we defined the relative rotation angles of the morphing panels, $\theta_i$ ($i=1$, $2$, and $3$), as the design variables. 
The primary objective of the optimization was to minimize the aerodynamic drag force, $F_{\mathrm{D}}$,  acting on the vehicle. 
The optimization problem can be formulated as
\begin{align}
&\mathrm{Minimize} \ F_{\mathrm{D}} (\theta_1, \theta_2, \theta_3) \nonumber \\ 
&\mathrm{subject\ to \ } (\theta_1, \theta_2, \theta_3) \in \Theta,
\end{align}
where $\Theta$ is the domain of feasible design variables, as described below. 

\begin{figure}%[H]
\centering
\includegraphics[width=0.48\textwidth]{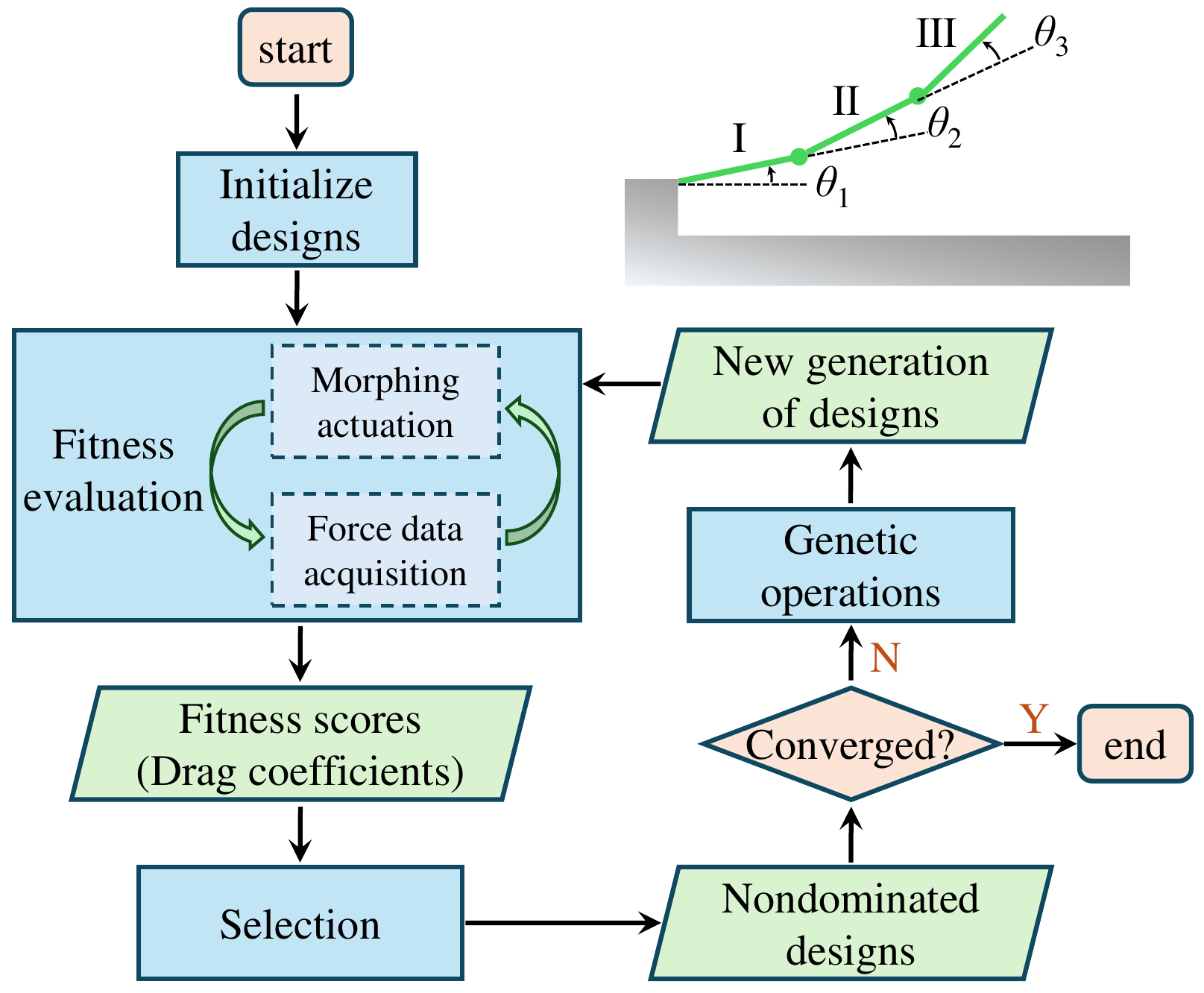}
\caption{
Flowchart of the key components of the GA-based autonomous optimization framework for morphing vehicles.   
The diagram on the top right corner illustrates the numbering of the panels and definitions of their rotation angle. 
}\label{fig:GA}
\end{figure}

At a given driving speed, optimization was achieved using a single-objective GA, as illustrated in Fig.~\ref{fig:GA}. 
The key components of the algorithm are outlined below. 

\noindent
\textit{Design space and constraints.} 
Each feasible morphing shape was described by the set of three design parameters, $\theta_i$ ($i=1$, $2$, and $3$), within their respective admissible range. 
Each design parameter was discretized into 64 increments, corresponding to resolutions of $0.5156^{\circ}$, $0.7031^{\circ}$, $0.7031^{\circ}$ for $\theta_1$, $\theta_2$, and $\theta_3$, respectively. 
The resulting design space consisted of a total of $274,625$ feasible morphing shapes. 

Constraints were imposed on the design parameters to ensure that the morphing structure maintained a clearance of $0.5\,\mathrm{cm}$ from the bed of the truck model to avoid collision. 
Additionally, the morphing structure was permitted to extend up to  $2.0\,\mathrm{cm}$ above the roof. 
While this upward deformation would intuitively lead to an increase in drag, we considered this possibility to allow for the exploration of unconventional designs that might lead to unexpected aerodynamic benefits. 
The set of all design parameters that reside within their admissible ranges and satisfy the constraints is denoted by $\Theta$. 

\noindent
\textit{Initialization.} 
The optimization process was initialized  with a population of randomly generated morphing shapes. 
A large initial population size of 50 was selected to ensure a diverse representation of potential designs.

\noindent
\textit{Fitness evaluation.}
To evaluate the efficiency of morphing, the fitness of each morphing strategy could be scored by the vehicle's fuel consumption in a realistic setting. 
Alternatively, the fuel economy could be indirectly estimated by measuring the aerodynamic drag force, $F_{\mathrm{D}}$, using force sensors. 
In this study, $F_{\mathrm{D}}$ was quantified in a controlled laboratory setting using a load transducer in the wind tunnel.
To account for the long-term effects of morphing while neglecting transient fluctuations caused by turbulence in the upstream flow and vortex shedding, drag force was measured and averaged over $10\,\mathrm{s}$, denoted as $\overline{F_{\mathrm{D}}}$. 
The detailed experimental procedure is described in the subsequent section.

\noindent
\textit{Selection.}
Upon evaluating the fitness of all morphing shapes in each generation, the optimal designs were selected based on their $\overline{F_{\mathrm{D}}}$ values. 
Here, we designated the four designs with the lowest $\overline{F_{\mathrm{D}}}$ as the ``elite'' individuals.

\noindent
\textit{Production.}
The elite individuals from each generation were used to generate offspring of morphing shapes for the subsequent generation. 
To this end, the design parameters, $\theta_i$ ($i=1$, $2$, and $3$), of the elites were encoded as strings of binary digits. 
Adopting the terminologies in biology, these binary representations were considered ``genes'', and the combination of all genes formed ``chromosomes''. 
Several standard genetic operations were performed on the elites' chromosomes to produce offspring. 
The elites were randomly grouped into pairs of parents, and four crossover points were randomly selected on their chromosomes to divide them into five segments. 
Each offspring inherited a segment of chromosomes from either parent with equal probability.

Once each offspring was produced, mutation was applied on their genes by flipping their binary digits at a $5\%$ probability. 
Offspring were discarded if they became inadmissible following crossover and mutation. 
The surviving offspring were combined with the elite individuals to form the next generation of designs. 
The production process was repeated until the population size of the next generation reached 20.

\noindent
\textit{Convergence.}
The iteration process depicted in Fig.~\ref{fig:GA} was repeated until either convergence was achieved or the designs had evolved through 50 generations.
Convergence was attained if no new elite designs was produced for five consecutive generations. 
Upon convergence, the morphing shape that minimized the drag force was identified as the optimal strategy.

This GA-based optimization framework was programmed in MATLAB. 
A similar program has been successfully implemented in our prior work to optimize the aerodynamic efficiency of highway infrastructure designs  \citep{Zhang21JIS}.

\subsection{Experimental setup and procedure}\label{sec:method:exp}

\begin{figure}%[H]
\centering
\includegraphics[width=0.48\textwidth]{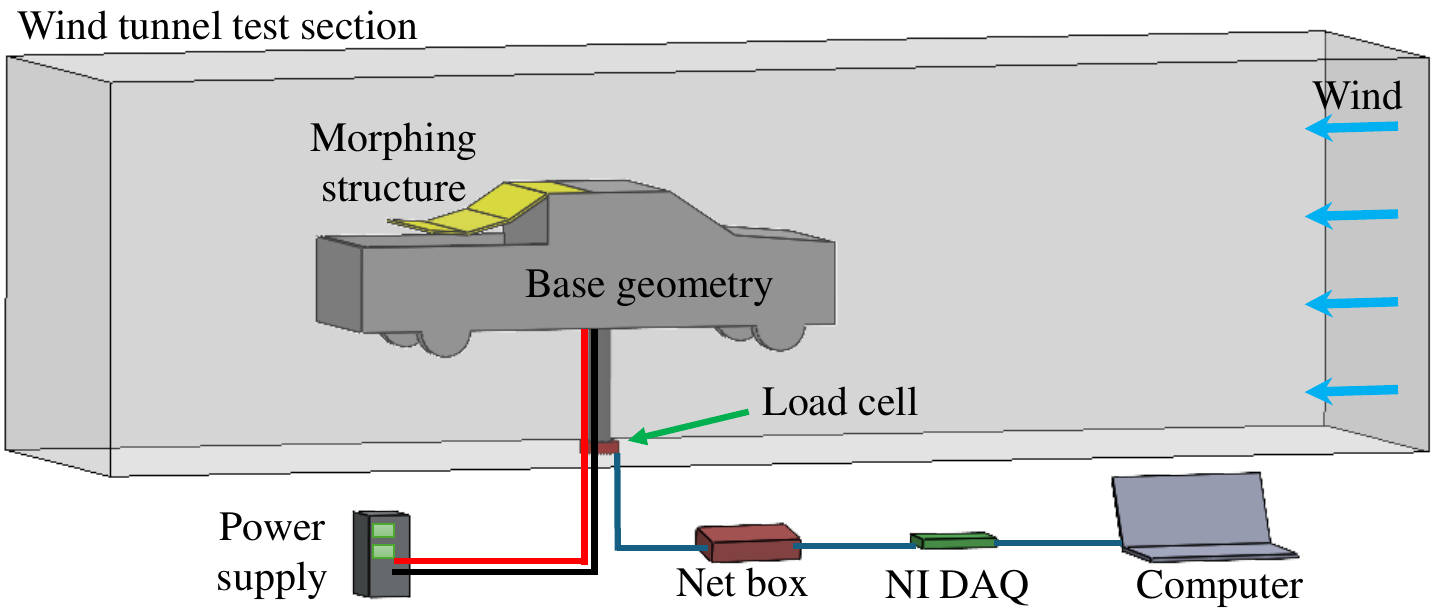}
\caption{ 
Schematic of the setup for the experiments in the wind tunnel
}\label{fig:exp_setup}
\end{figure}

To validate the performance of the proposed morphing vehicle prototype and optimization algorithm, a series of experiments were performed in a large low-speed wind tunnel (LSWT) facility. 
The LSWT was powered by a 75-horsepower motor fan capable of generating wind speeds up to $15\,\mathrm{m/s}$. 
It featured a $1.25\,\mathrm{m} \times 1.25\,\mathrm{m} \times 3.00\,\mathrm{m}$ (height$\times$width$\times$length) test section, which was equipped with a pitot tube for airflow speed characterization. 
%This platform is ideal for experiments on reduced-scale morphing truck models, capable of reproducing realistic aerodynamic conditions that closely mimic the field settings. 
The LSWT allowed for manual control of the wind speed by tuning the rotational frequency of its motor fan. 
Prior to our experiments, a series of calibration measurements were performed to establish a relationship between the wind speed and the rotational frequency of the motor fan.

The LSWT was equipped with a load quantification system, consisting of a load transducer (ATI Industrial Automation; model mini85 SI-1900-80) installed at the bottom of the test section for three-axes force and torque sensing, a net box (ATI Industrial Automation; model FTIFPS1) for processing and transmitting the transducer's readings, and a data acquisition (DAQ) board (National Instruments; model USB-6210) for data collection. 
The transducer could afford a load capacity of $\pm 1900\,\mathrm{N}$ along the axial direction and a maximum acquisition frequency of $10,000\,\mathrm{Hz}$. 
Preliminary tests confirmed that the system could resolve drag force variations smaller than $10^{-4}\,\mathrm{N}$; this level of precision is sufficient for our experiments. 
In this study, all data were collected at a sampling rate of $600\,\mathrm{Hz}$ using the DAQ and were saved on a computer via MATLAB's ``Data Acquisition Toolbox Support Package for National Instruments NI-DAQmx Devices''.

The proposed morphing vehicle prototype was installed in the LSWT to characterize its aerodynamic performance, as illustrated in Fig.~\ref{fig:exp_setup}. 
The vehicle model had a blockage ratio of 3.4\%, which was sufficiently small to minimize blocking effects.  
The vehicle model was mounted on the load transducer via a 3D-printed support positioned at its center of mass. 
Prior to each experimental trial, the motor fan was turned on for a 10-minute warm-up period to ensure steady-state aerodynamic conditions prior to data acquisition. 
All experiments were performed at a temperature between $10^{\circ}\mathrm{C}$ and $15^{\circ}\mathrm{C}$.

Two experiments were conducted in the LSWT to validate the performance of the proposed morphing vehicle prototype and the optimization algorithm. 
The first experiment (Exp 1) aimed to elucidate the impact of the morphing structure on the vehicle's aerodynamic performance. 
To this end, the drag forces were analyzed for two configurations: i) the base vehicle geometry without the morphing structure, and ii) the morphing vehicle with the structure in the neutral configuration (see Fig.~\ref{fig:validate_shapes}). 
A calibration procedure was performed for each trail. 
Specifically, forces were first acquired under wind-off condition and was treated as the reference force. 
The reference force was subtracted from the forces recorded under the wind-on conditions to obtain the net aerodynamic force induced by the airflow. 
Drag forces for each configuration were recorded at four upstream wind speeds: $U=5.79$, $7.33$, $8.65$, and $10.13\,\mathrm{m/s}$. 
Using the vehicle model's length of $L=78.11\,\mathrm{cm}$ as the characteristic length and the kinematic viscosity of air of $\nu = 1.48 \times 10^{-5}\,\mathrm{m^2/s}$ at $15^{\circ}\mathrm{C}$, the corresponding Reynolds numbers ($\Rey = L U / \nu$) were estimated to range from $3.1\times10^5$ to $5.3\times10^5$. 
Five repeated trials were conducted for each configuration and wind speed, with drag forces recorded for $30\,\mathrm{s}$ at a sampling rate of $600\,\mathrm{Hz}$ during each trial.

Building upon the preliminary understanding of the morphing structure's role, our second experiment (Exp 2) focused on evaluating the efficiency of the autonomous control algorithm in identifying the optimal structural shape. 
This experiment was designed based on the ``hardware-in-the-loop'' approach, where the wind tunnel facility was leveraged to evaluate the fitness of the structural morphing shapes, and the GA-based optimization algorithm described in Sec.~\ref{sec:method:algorithm} was implemented to iteratively search for the elite morphing strategy. 
Our approach was similar to the experimental scheme employed in Ref.~\citep{Boria09}, where the wind tunnel measurements were seamlessly integrated with GA for the shape optimization of a morphing wing. 
To account for potential drift of the load transducer signal over extended periods, the drag force was recalibrated before each generation. 
Unlike Exp 1, the drag force in Exp 2 was calibrated against the aerodynamic drag experienced by the neutral configuration, thereby avoiding the need to repeatedly turn the LSWT on and off and reducing overall experimental time. 
Consequently, the transducer measurements quantified the drag variation induced by the morphing structure, denoted as $\Delta F_{\mathrm{D}}$. 
Due to its extended duration of this experiment, measurements were performed only at a wind speed of $U=7.33\,\mathrm{m/s}$.

To further validate and visualize the effect of morphing, additional experimental trials were conducted to investigate the elite morphing shapes. 
In each trial, the morphing structure was initially maintained at the neutral configuration for $10\,\mathrm{s}$ before transitioning to the elite shape, which was held for an additional $15\,\mathrm{s}$. 
The drag force was recorded throughout the entire dynamic morphing event. 
Four repeated trials were performed for each elite morphing strategy, and all trials were performed at a wind speed of $U=7.33\,\mathrm{m/s}$. 
Actuation of the morphing structure via Arduino and force data acquisition via DAQ were accomplished using two MATLAB sessions, which were synchronized  through TCP/IP communication.

\begin{figure*}%[H]
\centering 
\subfigure[]{\includegraphics[width=0.45\linewidth]{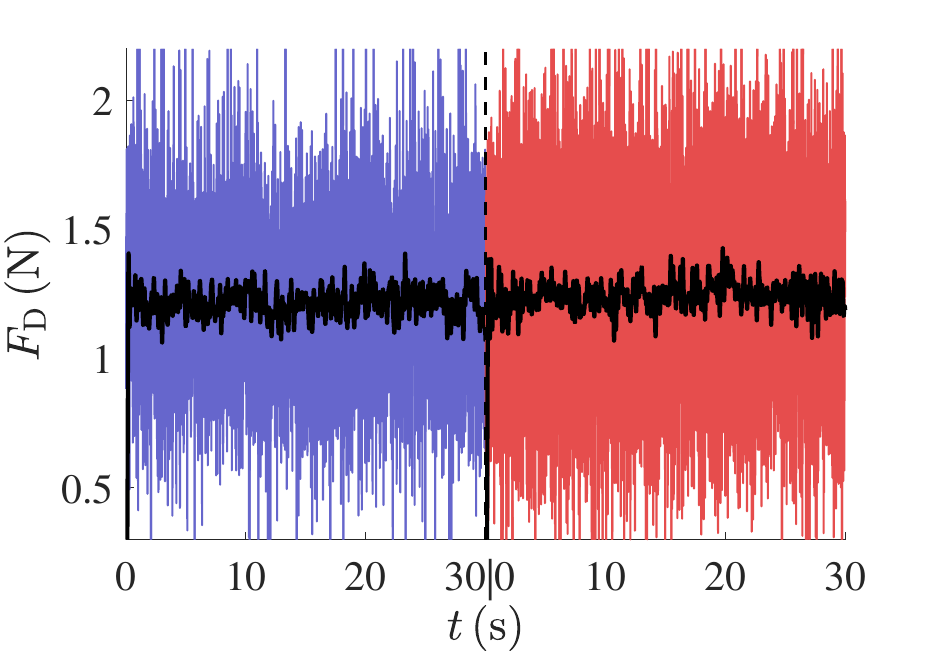}}
\subfigure[]{\includegraphics[width=0.45\linewidth]{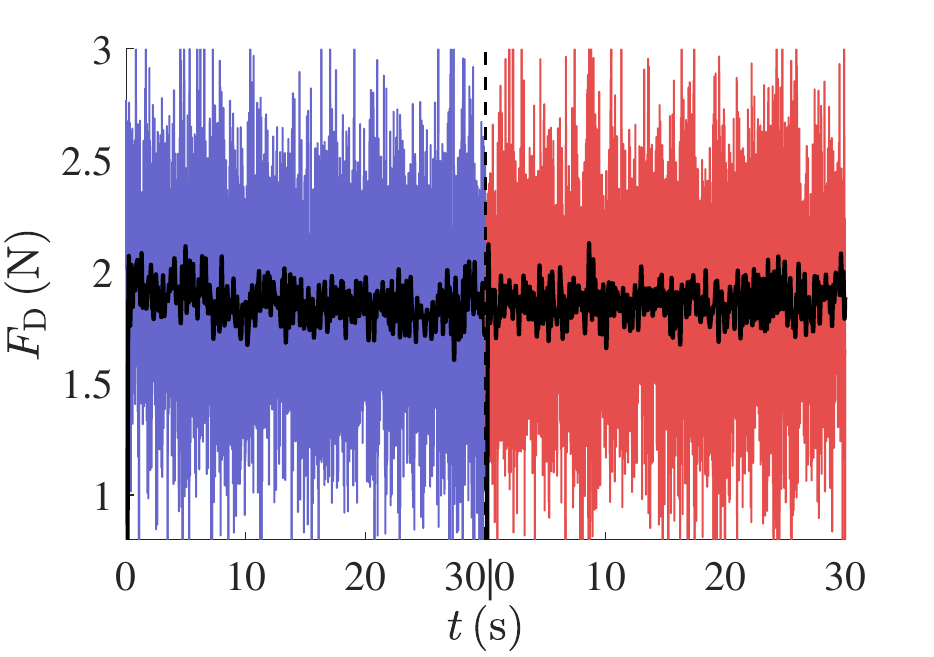}}\vspace{-0.8\baselineskip}
\subfigure[]{\includegraphics[width=0.45\linewidth]{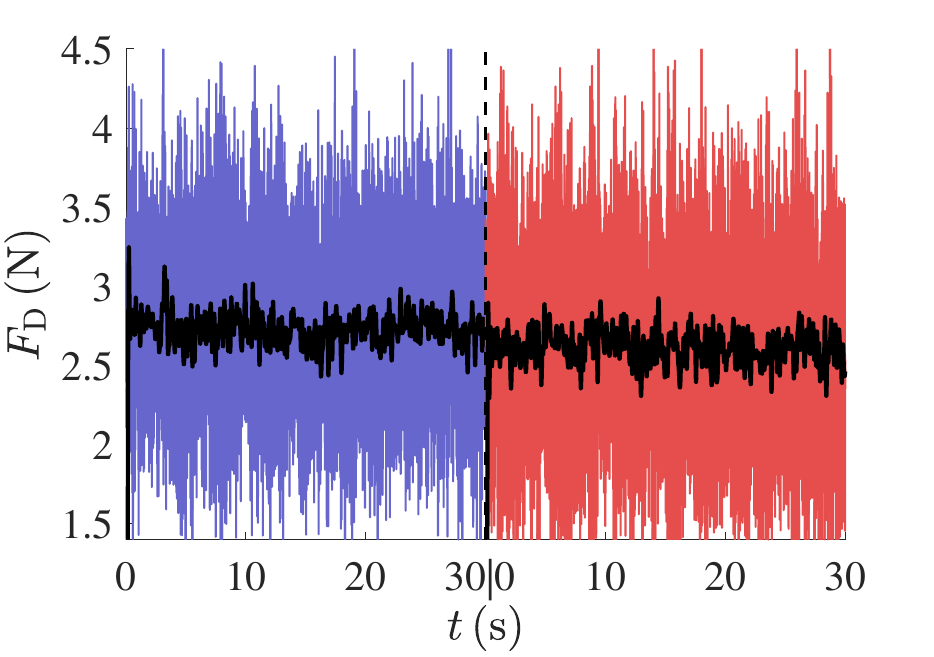}}
\subfigure[]{\includegraphics[width=0.45\linewidth]{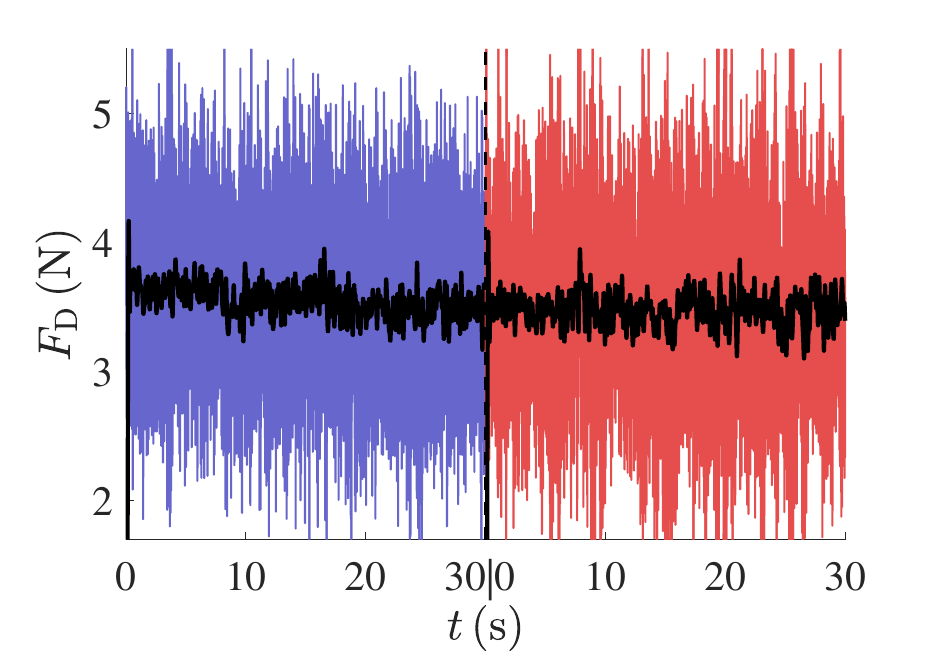}}%\vspace{-0.8\baselineskip}
\caption{
\label{fig:flat-drag} 
Time histories of the drag force for the base vehicle without structure (blue) and the morphing vehicle prototype at the neutral configuration (red). 
Colored (blue and red) curves are raw force data, while solid black curves are data processed by the low-pass filter with a cutoff frequency of $5\,\mathrm{Hz}$.
}
\end{figure*}

\subsection{Hypotheses and analysis methods}\label{sec:method:analysis}

We performed two-sample t-tests to compare the time-histories of drag force data.  
In Exp 1, we hypothesized that $F_{\mathrm{D}}$ on the morphing vehicle at the neutral configuration was greater than $F_{\mathrm{D}}$ on the base vehicle geometry. 
Similarly, in Exp 2, we hypothesized that $F_{\mathrm{D}}$ on the elite morphing shape was greater than $F_{\mathrm{D}}$ on the neutral configuration. 
in both cases, the null hypothesis was that the mean drag forces on the two configurations were independently sampled from normal distributions with equal means and equal but unknown variances. 
All t-tests were performed at the significance level of $0.01$.

The raw drag force data acquired in the wind tunnel contained substantial high-frequency noise, likely due to vortex-induced vibrations of the structures and moderate levels of turbulence. 
To facilitate the visualization of mean drag forces experienced by the vehicle, the noisy raw data were processed using a low-pass filter with a cutoff frequency of $5\,\mathrm{Hz}$. 
We emphasize that low-pass filters were employed for visualization purposes only; all statistical analyses were performed on unfiltered raw force data.

\section{Results and discussion}\label{sec:results}

\subsection{Impact of the morphing structure}\label{results:flat}

To establish a baseline understanding of the aerodynamic characteristics of the vehicle models, we quantified the drag force on the base vehicle and the morphing vehicle in the neutral configuration, as shown in Table \ref{tab:drag}. 
As the upstream wind speed increased from  $U=5.79\,\mathrm{m/s}$ to $10.13\,\mathrm{m/s}$, the mean drag force ($\overline{F_{\mathrm{D}}}$) on the base vehicle model increased monotonically from $1.227\,\mathrm{N}$ to $3.551\,\mathrm{N}$. 
The corresponding drag coefficient, $C_{\mathrm{D}} = 2 \overline{F_{\mathrm{D}}} / \left( \rho_{\mathrm{a}} U^2 A  \right)$,  where the frontal area of the vehicle model was $A = 5.321\times 10^{-2} \,\mathrm{m^2}$ and the density of air was $\rho_{\mathrm{a}} = 1.225 \,\mathrm{kg/m^3}$ at $15^{\circ}\mathrm{C}$, varied within a narrow range between $1.062$ and $1.124$.

Comparisons of the drag forces between the base vehicle and the morphing vehicle in the neutral configuration pointed at a negligible impact of the morphing structure on aerodynamic drag, as evidenced in Fig.~\ref{fig:flat-drag}. 
While a marginal difference was observed at low wind speeds, a modest reduction in the drag force became evident at higher wind speeds.
Statistical analyses supported these findings, showing a significant decrease in drag force at $U=7.33$, $8.65$, and $10.13\,\mathrm{m/s}$ ($p<0.01$ in two-sample t-tests), while the null hypothesis was upheld for $U=5.79\,\mathrm{m/s}$.

\begin{table}
%\begin{center}
 \caption{\label{tab:drag} Mean drag forces ($\overline{F_{\mathrm{D}}}$) measured on the base vehicle model and morphing vehicle at the neutral configuration.}
 \begin{tabular}{ccc} %{c|c|c|c|c} % 
 \hline \hline
$U$ ($\mathrm{m/s}$)  &  Base vehicle ($\mathrm{N}$) & Neutral config.~($\mathrm{N}$) \\
 \hline 
 
5.79 & $1.227 \pm 0.263$ & $1.259 \pm  0.383$    \\
7.33 &  $1.901 \pm  0.334 $ & $1.888 \pm  0.353 $ \\
8.65 & $2.703 \pm  0.468$ & $2.621 \pm 0.513$ \\
10.13 & $3.551 \pm  0.586$ & $3.485 \pm  0.655$ \\

\hline \hline
\end{tabular}
%\end{center}
\end{table}

Despite the statistical significance, the drag force reduction achieved by the neutral configuration was minimal. 
The largest observed reduction occurred at $U = 8.65\,\mathrm{m/s}$, with a decrease of $0.082\,\mathrm{N}$, corresponding to only $3\%$ of the total drag force. 
These results highlight the necessity of employing an optimization algorithm to identify morphing structure shapes that can more effectively minimize aerodynamic drag.

\subsection{Optimal morphing strategies}\label{results:optimal}

\begin{figure}%[H]
\centering 
\subfigure[]{\includegraphics[width=0.95\linewidth]{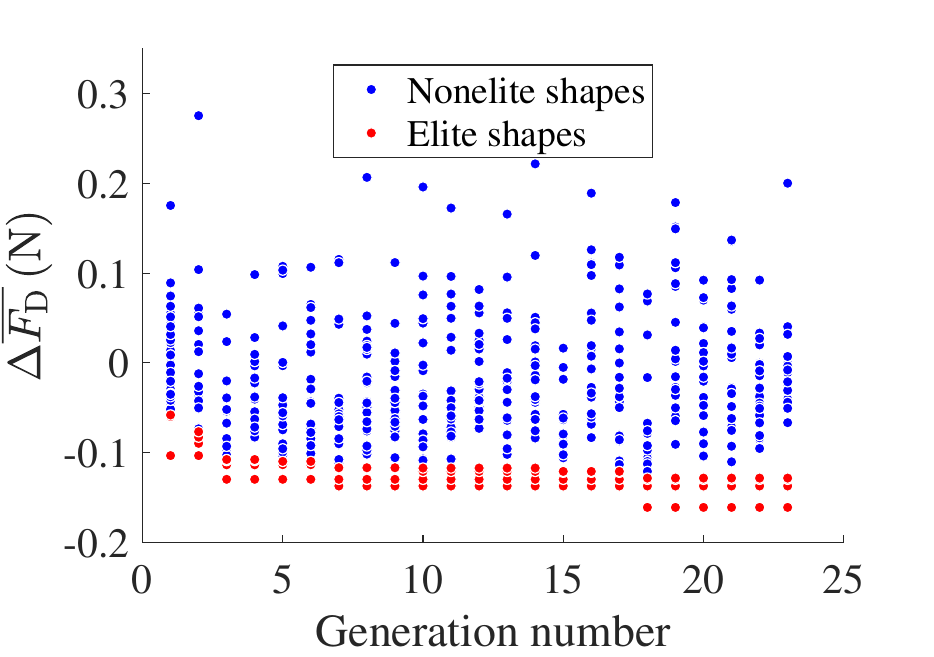}}\vspace{-0.8\baselineskip}
\subfigure[]{\includegraphics[width=0.95\linewidth]{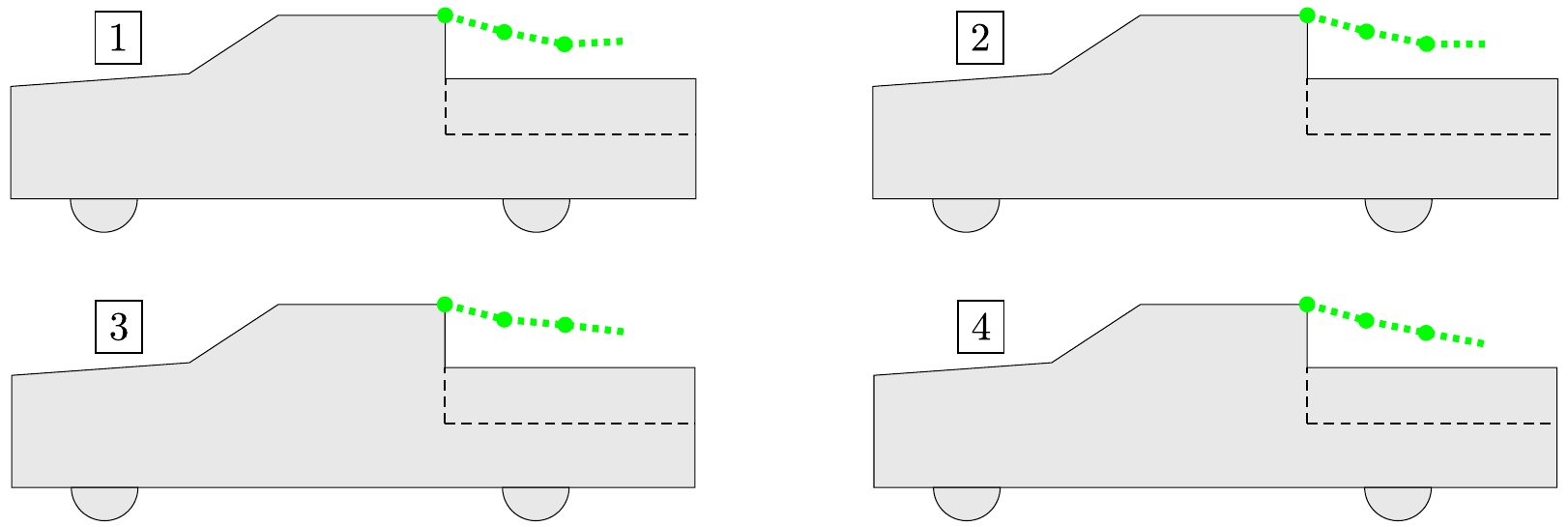}}
\caption{
\label{fig:optimal} 
(a) Evolution of the mean drag forces as a function of generation number. 
(b) Illustration of the four optimal morphing shapes identified by the GA, where cases labeled 1--4 attained $8.5\%$, $7.3\%$, $6.9\%$, and $6.8\%$ reductions in the mean drag force, respectively. 
}
\end{figure}

The GA-based optimization converged rapidly after 23 generations, exploring a total of 529 morphing shapes. 
The duration of the entire optimization process was approximately $2.5\,\mathrm{h}$.
Compared with CFD-based optimization, which often requires days or even weeks of computational time, this experimental approach proved significantly more efficient.

The optimization algorithm successfully identified the optimal morphing shapes. 
As shown in Fig.~\ref{fig:optimal}(a), four elite morphing strategies were selected in each generation, and their corresponding drag forces decreased progressively as a function of generation. 
Upon convergence, the best four morphing shapes were identified, as depicted in Fig.~\ref{fig:optimal}(b). 
The morphing shapes labeled 1--4 in Fig.~\ref{fig:optimal}(b) attained $8.5\%$, $7.3\%$, $6.9\%$, and $6.8\%$ reductions in $F_{\mathrm{D}}$, respectively,  compared with the vehicle in the neutral configuration.

As shown in Fig.~\ref{fig:optimal}(b), the four elite morphing strategies demonstrated several common features. 
Interestingly, a downward slope was observed at Panels \RNum{1} and \RNum{2} in all elites shapes. 
However, two distinct deformation strategies emerged for Panel \RNum{3} at the free end. 
For elites 1 and 2, Panel \RNum{3} maintained a nearly horizontal orientation, whereas in elites 3 and 4, it continued the downward slope observed in the other panels.

\begin{figure}[H]
\centering 
\subfigure[]{\includegraphics[width=0.95\linewidth]{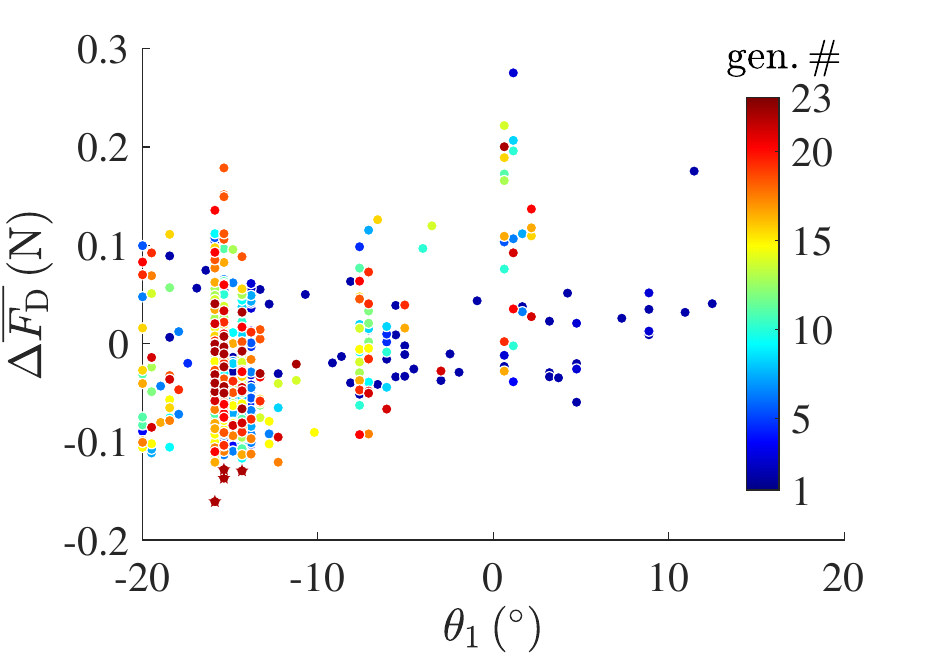}}\vspace{-0.8\baselineskip}
\subfigure[]{\includegraphics[width=0.95\linewidth]{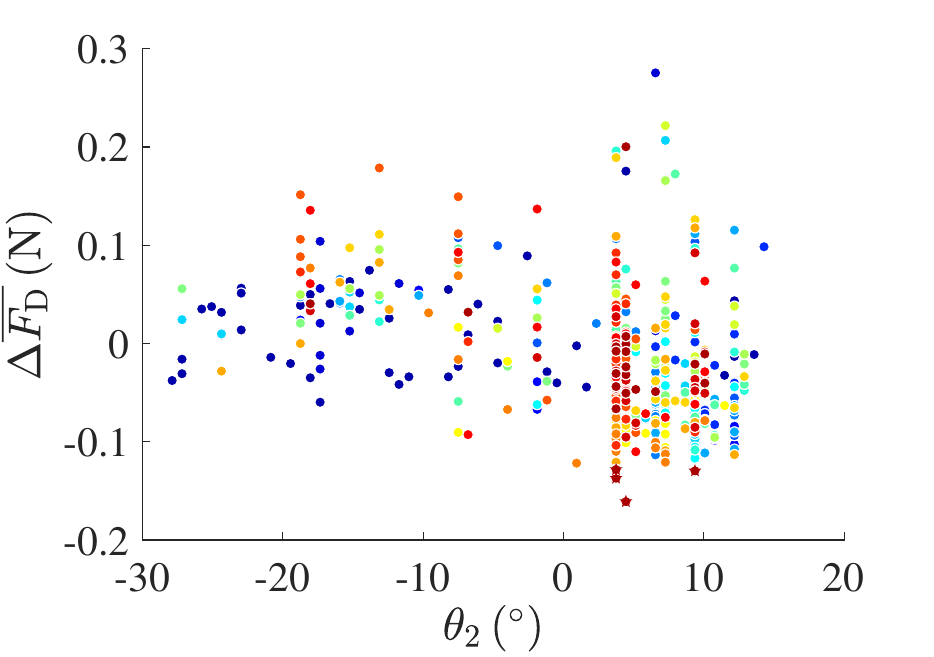}}\vspace{-0.8\baselineskip}
\subfigure[]{\includegraphics[width=0.95\linewidth]{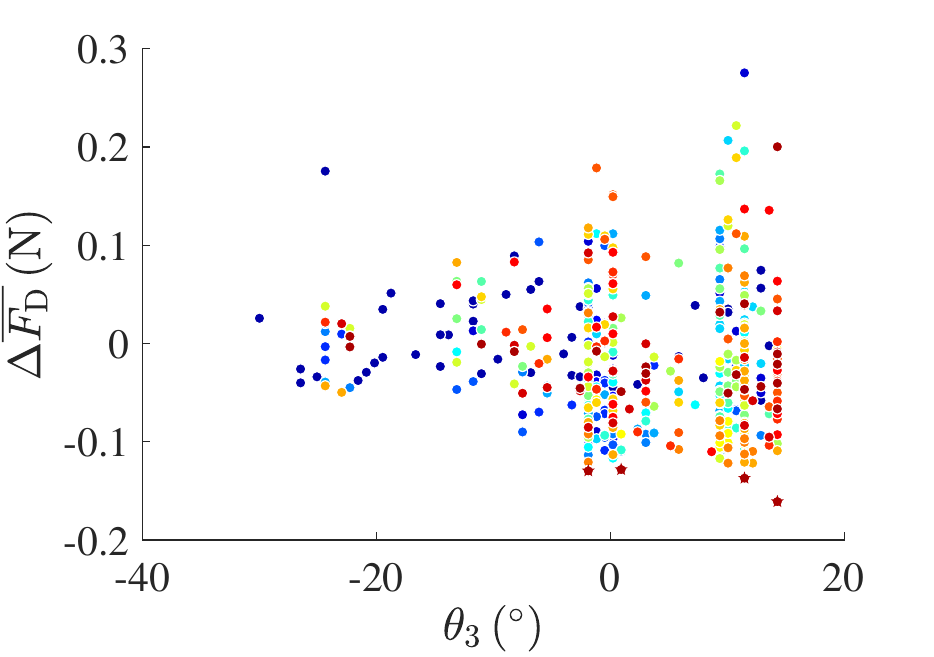}}
\caption{
\label{fig:theta} 
Evolution of the three design parameters as a function of generation number. 
Color represents the generation number.
The elite morphing parameters are labeled as stars. 
}
\end{figure}

Further inspections of the design parameters corroborated this observations. 
As shown in Fig.~\ref{fig:theta}, Panel \RNum{1} consistently adopted a similar angle of approximately $\theta_1 = 15^{\circ}$ across all elite morphing strategies.
Likewise, Panel \RNum{2} displayed a slight upward rotation  relative to Panel \RNum{1}, with $\theta_2$ ranging from $4^{\circ}$ to $9^{\circ}$. 
While Panel \RNum{3} in elites 1 and 2 exhibited a significant upward rotation ($11^{\circ} < \theta_3 < 14^{\circ}$), its rotation was minimal in elites 3 and 4 ($-2^{\circ} < \theta_3 < 1^{\circ}$).

The efficiency of the GA-based optimization framework was  further evidenced by the evolution of the three rotation angles, as depicted in Fig.~\ref{fig:theta}. 
While the rotation angles in the initial generation randomly spanned the entire admissible design space, they progressively clustered around the optimal values as the structural shape evolved. 
Notably, a small fraction of designs were still explored in later generations as a result of mutation, ensuring diversity in the design parameters that are critical for escaping local minima and achieving global optimization.

Although our experimental setup lacked flow visualization capabilities to directly investigate the drag reduction mechanism, previous work could provide insights into the underlying flow physics. 
The elite morphing shapes depicted in Fig.~\ref{fig:optimal} closely resemble the optimal structural shapes identified in our recent CFD simulations \citep{Kazemipour24}, where the morphing structure exhibited a downward slope near the base and gradually transitioned to a flatter orientation at the free end. 
Analysis of the velocity field in that study suggested that this structural shape streamlined the flow in the bed region and reduced the size of the circulating wake, thereby decreasing the drag force.

\begin{figure*}%[H]
\centering
\subfigure[]{\includegraphics[width=0.45\linewidth]{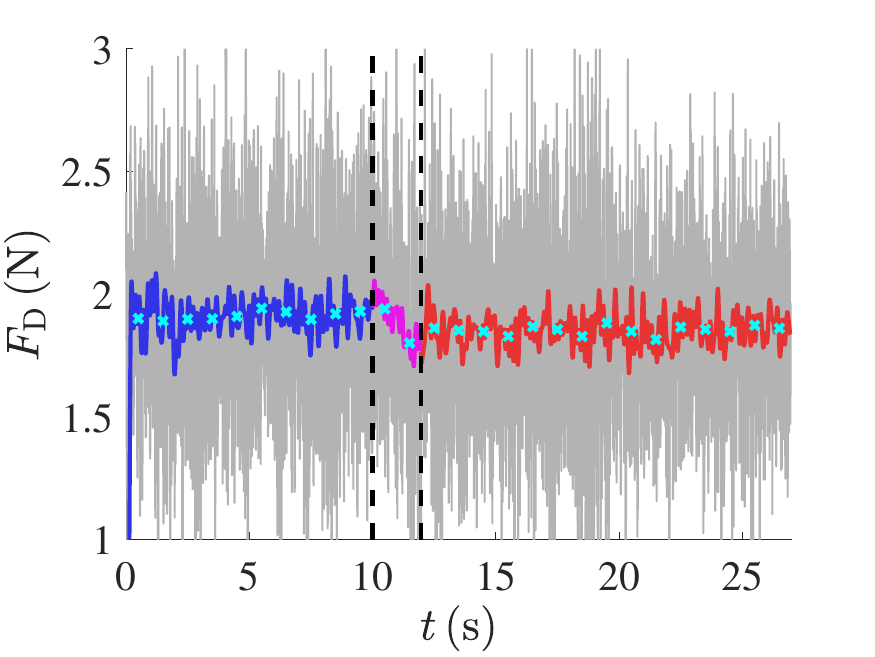}}
\subfigure[]{\includegraphics[width=0.45\linewidth]{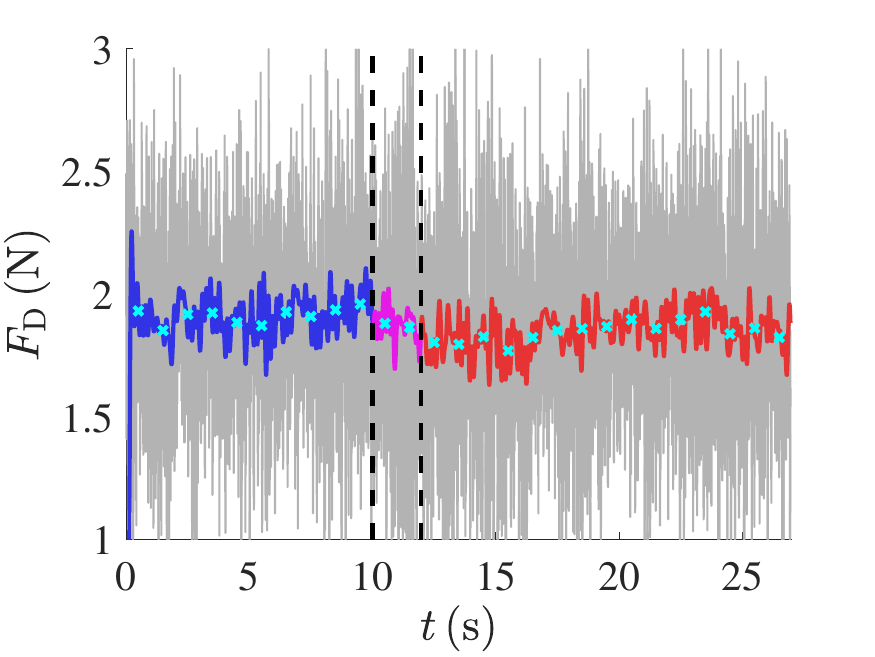}}\vspace{-0.8\baselineskip}
\subfigure[]{\includegraphics[width=0.45\linewidth]{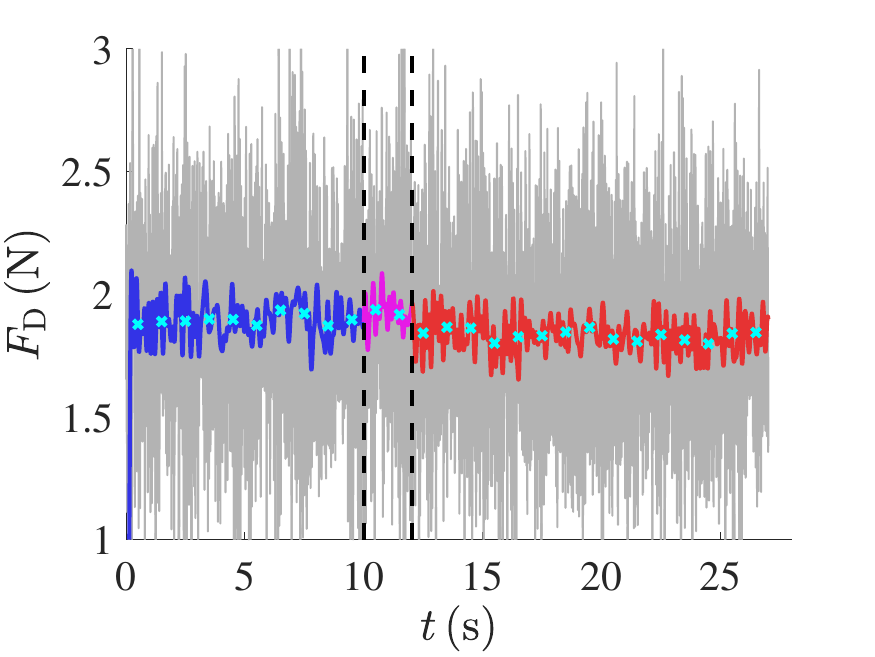}}
\subfigure[]{\includegraphics[width=0.45\linewidth]{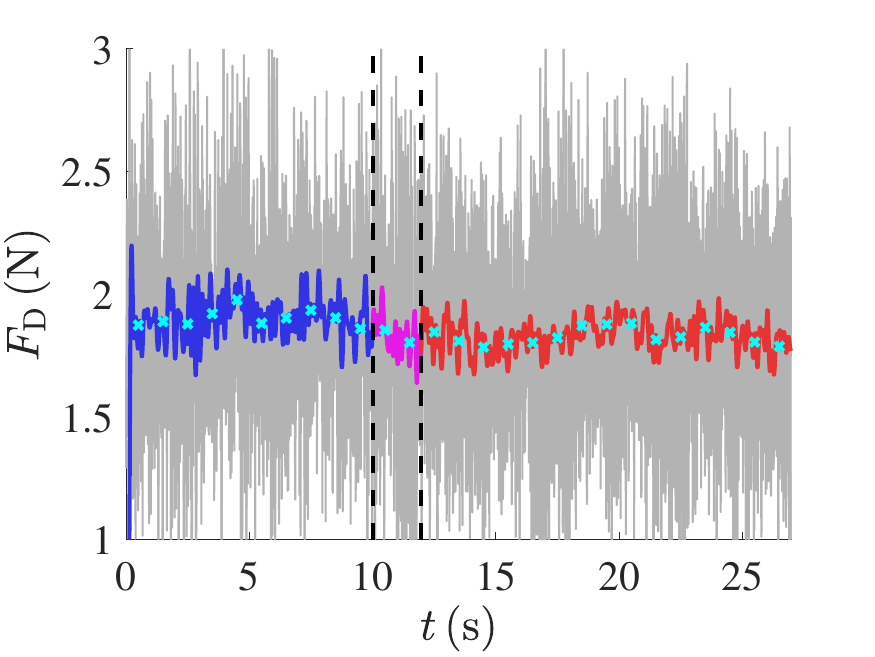}}
\caption{Time histories of the drag forces during dynamic deformation of the morphing structure from the neutral position to the elite shapes. 
(a) -- (d) correspond to the elite shapes 1--4 in Fig.~\ref{fig:optimal} (b), respectively. 
Light gray curves represent raw drag force data acquired by the load cell. 
Blue and red curves represent data processed by a low-pass filter with a cutoff frequency of $5\,\mathrm{Hz}$. 
Cyan crosses represent averaged force data over one-second time windows. 
Black dashed lines at approximately $t=10\,\mathrm{s}$ and $12\,\mathrm{s}$ mark the onset and end of the morphing event. 
}\label{fig:dynamic}
\end{figure*}

To further investigate the elite morphing strategies' capability of regulating aerodynamic drag, we inspected the dynamic variations in the drag force as the morphing structure transitioned from the  neutral configuration to the elite shape,  as shown in in Fig.~\ref{fig:dynamic}. 
To highlight the long-term variations in the drag force, denoised data using the low-pass filter and a one-second moving average window were overlaid in Fig.~\ref{fig:dynamic}. 
Across all elite strategies,  the drag force was notably lower on the elite configuration than on the neutral configuration. 
An evident decrease in the drag force was noted as the structure transitioned from the neutral configuration to the elite shapes between $t=10\,\mathrm{s}$ and $12\,\mathrm{s}$. 
Remarkably, statistical analyses confirmed a significant  reduction in drag from the neutral configuration to the optimal shapes for all four elites and across all repeated trials ($p<0.01$ in two-sample t-tests).

\subsection{Design of full-size morphing vehicles}\label{results:fullsize}

The scaled-down morphing vehicle prototype presented in this work offers valuable insights into the design of full-size morphing vehicles. 
The primary challenge at full scale is the increased demand for torque and power to accomplish morphing. 
Simply scaling up the dimensions of the PLA panels by a factor of eight would substantially increase their weight, exceeding the practical torque capacity of standard actuators. 
To ameliorate this issue, lightweight and high strength materials, such as fiber-reinforced polymers, could be adopted for fabricating the morphing panels. 
Potential candidates include carbon fiber-reinforced polymers (CFRPs), glass fiber-reinforced polymers (GFRPs), and aramid fiber-reinforced polymers (AFRPs), which have been widely used in aerospace and automotive industries \cite{Ozkan20}.  
The mechanical properties of some fiber-reinforced polymers are presented in Table \ref{tab:materials}. 
To assess their suitability for morphing structure design, we perform an order-of-magnitude analysis to estimate the panel dimensions and corresponding actuation torque requirements.

\begin{table*}
\begin{footnotesize}
\begin{center}
 \caption{\label{tab:materials} Candidate materials for fabricating full-size morphing structures}
 \begin{tabular}{cccccc} %{c|c|c|c|c} % 
 \hline \hline
Material  &  E ($\mathrm{GPa}$) & $\sigma_{\mathrm{u}}$ ($\mathrm{MPa}$) & $\rho$ ($\mathrm{kg/m^3}$) & $2h_{\mathrm{min}}$ ($\mathrm{mm}$) & Actuation torque ($\mathrm{N} \cdot \mathrm{m}$) \\
 \hline 
 
CFRP (70\% fibers in epoxy) \cite{Subtech} & $181$ & $1500$  & $1.6\times10^3$ & $3.5$ & $149.7$ \\
CFRP (80\% fibers in polyetherimide) \cite{Sharma10} &  $130 $ & $517 $ & $1.25\times10^3$ & $3.7$ & $122.0$ \\
GFRP (35\% fibers in epoxy) \cite{Megahed21} & $34$ & $157$ & $1.47\times10^3$ & $7.8$ & $305.5$ \\
AFRP (60\% fibers in epoxy) \cite{Edwards98} & $75$ & $1400$ & $1.4\times10^3$ & $5.1$ & $190.4$ \\

\hline \hline
\end{tabular}
\end{center}
\end{footnotesize}
\end{table*}

To maintain geometric similarity with the proposed scaled-down prototype, we extend the length and width of the panels by a factor of eight. 
On the other hand, the thickness of the morphing panels, $2h$, could be carefully adjusted to minimize the total weight, avoid material failure, and maintain sufficient rigidity. 
For fully extended morphing panels illustrated in Fig.~\ref{fig:optimal}(b), the mechanical stress primarily arises from their own weight. 
Additional aerodynamic forces could also induce stresses within the panels. 
Specifically, on a moving vehicle, the morphing structure would induce a faster airflow over its top surface and a slower circulating flow underneath, potentially generating a net lift force. 
Although this lift force may partially offset the structural weight, we adopted a conservative estimation by neglecting its contribution.

%\noindent

Assuming a rectangular cross-sectional shape, the maximum normal stress occurs at the clamped end of the panel at its top and bottom fibers. 
This stress magnitude can be expressed as $\sigma_{\mathrm{max}} = 3 \rho g l^2 / 2h$, where $g=9.8\,\mathrm{m/s^2}$ is the gravitational acceleration, $ \rho$ is the material density, and $l=1.68 \mathrm{m}$ is the total length of all morphing panels. 
To prevent material failure, this stress must remain within the material's ultimate tensile strength, $\sigma_{\mathrm{max}} \leq \sigma_{\mathrm{u}}$.
As a result, the half-thickness of the panel should satisfy the following condition:
\begin{equation}\label{eq:h-stress}
h \geq \frac{3 \rho g l^2}{2 \sigma_{\mathrm{u}}}.
\end{equation}

Similarly, the elastic deflection of the panels under their own weight should be sufficiently small to preserve the desired morphing shapes. 
For morphing panels fully extended in the horizontal orientation, the maximum deflection, $\delta_{\mathrm{max}}$, occurs at the free end.
Approximating the panels as cantilever beams subjected to their own distributed weight, the maximum deflection can be estimated as $\delta_{\mathrm{max}} = 3 \rho g l^4 / (8Eh^2)$, where $E$ is the Young's modulus. 
To ensure sufficient structural rigidity, we define a permissible relative deflection, $\gamma$, such that $\delta_{\mathrm{max}}/l \leq \gamma$.
Consequently, the half-thickness must satisfy
\begin{equation}\label{eq:h-deflect}
h \geq \sqrt{\frac{3 \rho g l^3}{8 E \gamma }}.
\end{equation}
The minimum half-thickness $h_{\mathrm{min}}$ is determined as the minimum value that satisfies both Eqs.~\eqref{eq:h-stress} and \eqref{eq:h-deflect}. 
The computed values of $h_{\mathrm{min}}$ for the candidate fiber-reinforce materials, assuming a permissible relative deflection of $\gamma=5\%$, are summarized in Table \ref{tab:materials}.

The morphing panels fabricated from the materials presented in Table \ref{tab:materials} could be linked and actuated by commercial  limited-angle torque motors. 
The actuation torque required for morphing can be estimated based on the bending moment generated by the panels' weight at their clamped end as $\tau = \rho g l^2 w h$, where $w=1.93 \mathrm{m}$ is the width of the full-size morphing panels. 
As presented in Table \ref{tab:materials}, the actuation torque $\tau$ ranges between $100\,\mathrm{N} \cdot \mathrm{m}$ and approximately $300\,\mathrm{N} \cdot \mathrm{m}$, which falls within the capability of commercially available electric motors. 
These motors can be commanded by onboard microcontrollers programmed to execute the autonomous learning algorithm to determine the optimal morphing strategy. 
All electronic components associated with the morphing structures operate at low voltages and thus can be directly powered by the vehicle's onboard battery.

\subsection{Estimation of energy savings and reduction in green-house gas emissions}

Wind tunnel testing of reduced-scale models is a cost-effective method for assessing the aerodynamic performance of full-size vehicles \cite{Meinert16}. 
Although matching the aerodynamic conditions between the wind tunnel tests and realistic road environments remains challenging, analyses of scaled-down models can afford valuable insights into the design of full-size vehicles. 
Beyond a critical $\Rey$, the flow within the boundary layers and the wake becomes fully turbulent, and further increases in $\Rey$ do not significantly alter the drag or lift coefficients. 
Previous wind tunnel experiments on reduced-scale models have demonstrated $\Rey$-independence of the drag coefficient for $\Rey>1\times10^5$ on a heavy truck model \cite{Bayindirli16} and for $\Rey>5\times10^5$ on a Winsor body \cite{Howell23}. 
Our measurements align with these findings: while the $\Rey$ in this study ranged from $3.1\times10^5$ to $5.3\times10^5$, corresponding to a 75\% increase, the value of $C_{\mathrm{D}}$ varied by only 6\%. 
Consequently, it is reasonable to posit that  a similar maximum drag reduction of approximately 8.5\% could be realized by full-size morphing vehicles operating under realistic road conditions.

Reductions in aerodynamic drag force could provide considerable economic and environmental benefits. 
Here, we perform an order-of-magnitude estimation of the potential energy cost savings and GHG emission reductions for gasoline-powered pickup trucks. 
Assuming a gasoline price of $\$0.66/\mathrm{l}$ ($\$2.5 / \mathrm{gal}$) and a representative pickup truck fuel efficiency of $8.5\,\mathrm{km/l}$ ($20\,\mathrm{miles/gal}$), the $8.5\%$ drag reduction results in fuel cost savings of $\$6.6\times10^{-3}/\mathrm{km}$ ($\$1.1\times10^{-2}/\mathrm{mile}$). 
Given the average annual distance of  $18,215\,\mathrm{km}$ ($11318\,\mathrm{miles}$) traveled by pickup trucks \cite{AFDC}, this corresponds to an annual saving of approximately $\$120$.

While drag reductions results in energy savings,  these benefits must be weighed against the energy consumed by the morphing process itself.
To estimate the electric energy required for each morphing event, we assume that the panels are initially aligned vertically along the rear window and are subsequently morphed to the horizontal neutral configuration shown in Fig.~\ref{fig:validate_shapes}. 
Although actual morphing often require smaller modifications to the structural shape, this scenario provides a conservative upper-bound estimation of the energy expenditures associated with morphing. 
Assuming a driving speed of $20\,\mathrm{m/s}$ ($45\,\mathrm{mph}$) and a motor efficiency of $50\%$, the vehicle would need to travel less than $20\,\mathrm{m}$ to recover the morphing energy cost through aerodynamic savings for all candidate materials listed in Table \ref{tab:materials}. 
Such rapid energy recovery suggests that morphing vehicles can achieve improved aerodynamic efficiency not only during extended steady-speed highway driving, but also during frequent speed variations typical  in residential and urban settings.

To evaluate the environmental benefits of morphing vehicles, we estimate the potential reductions in carbon dioxide (CO$_2$) emissions. 
Based on gasoline's CO$_2$ content of $2.35\,\mathrm{kg/l}$ ($8.88\,\mathrm{kg/gal}$) \citep{EPA_emissions}, an $8.5\%$ drag reduction corresponds to an  emission rate of $2.3\times 10^{-2}\,\mathrm{kg/km}$ ($3.8\times 10^{-2}\,\mathrm{kg/mile}$). 
Based on the average annual travel distance of pickup trucks, this results in an estimated total annual CO$_2$ emission reduction of $427\,\mathrm{kg}$.

\section{Conclusion}\label{sec:conclude}

Road vehicles are responsible for significant levels of greenhouse gas emissions. 
Despite continued efforts to improve their fuel economy, the suboptimal exterior shapes of medium and heavy-duty vehicles continue to impede their aerodynamic efficiency. 
To address this issue, we proposed a morphing vehicle concept capable of actively interacting with the aerodynamic environment toward enhanced fuel economy. 
Morphing was accomplished by retrofitting a deformable structure consisting of three hinged plastic panels. 
Deformation was enabled by servo motors that can precisely control the relative rotation of the panels. 
The feasibility of this design concept was demonstrated on a reduced-scale morphing vehicle prototype. 
Contrary to exiting morphing vehicle designs that rely on intrusive modifications to the vehicles' geometry, the proposed morphing structure could be readily retrofitted to existing vehicle models, thereby significantly reducing design costs and production time.

The proposed mechanical design was integrated with a GA-based optimization algorithm that could autonomously identify the  structural shape that minimized aerodynamic drag.  
The efficiency of the design and the optimization algorithm was validated through an extensive experimental campaign in the LSWT. 
The optimization framework demonstrated excellent efficiency in autonomously identifying the optimal structural shape, whereby the entire optimization process concluded in under $2.5\,\mathrm{h}$ after exploring over 500 feasible shapes. 
Our experiments highlighted the feasibility of real-time shape optimization under conditions representative of realistic road environments.

The optimal morphing shape identified through GA-based optimization elicited an $8.5\%$ reduction in the mean drag force, corresponding to an equivalent increase in the vehicle's driving range. 
The accuracy of the optimization was validated by analyzing the dynamic variation in the drag force as the morphing structure transitioned from the neutral configuration to the elite shapes. 
During this dynamic process transition, we observed a clear and consistent decrease in the drag force. 
Statistical analyses confirmed a significant reduction in drag force from the neutral configuration to the elite morphing shapes.

Wind tunnel experiments on the scaled-down model provided critical insights into the design guidelines for full-size morphing vehicles.
We identified a list of lightweight and high-strength materials as viable candidates to construct morphing panels on full-size vehicles. 
Through an order-of-magnitude analysis, we estimated that the morphing structures could yield an appreciable fuel cost saving of $\$120$ and  reductions in CO$_2$ emissions by $427\,\mathrm{kg}$ annually for a typical pickup truck.

Despite the success of the GA-based optimization framework, it can be further enhanced to tackle the complex aerodynamic environments often encountered in the field. 
Transient aerodynamic events, such as variations in driving speed and unsteady wind gusts, demand continuous morphing strategies that dynamically adapt to the changing environment. 
A naive solution would be to perform GA-based optimization \textit{a priori} for all possible aerodynamic conditions; however, the associated costs can become prohibitively high, especially for aerodynamic conditions influenced by multiple input factors.  
A viable approach to tackle this challenge is through parametric GA, such as the predictive parametric Pareto GA proposed by \cite{Galvan15}. 
Within a parametric GA framework, both the controllable design variables and  uncontrollable input aerodynamic parameters  are incorporated into the optimization process. 
This framework effectively establishes a continuous representation of the dominator attributes, enabling instantaneous predictions of the optimal morphing strategy based on the transient aerodynamic environment \citep{Rosen20}.  
This parametric GA-based optimization framework has been successfully implemented in our recent work for optimizing morphing structures across a range of driving speeds in a computational environment \citep{Kazemipour24}. 
Future work should focus on integrating parametric GA-based optimization in the proposed design concept  to establish adaptive morphing vehicles capable of operating in dynamic aerodynamic environments.

Several other research directions will be pursued in the future.  
In this study, the optimal morphing shape only achieved a moderate drag reduction of 8.5\%, likely due to the use of rigid panels that produced non-smooth shapes.
Future work should explore smooth, extensible morphing structures to further streamline the airflow and enhance aerodynamic efficiency. 
Additional morphing structures could be retrofitted to the sides and wake of the vehicle to exploit the full aerodynamic potential of shape morphing. 
Likewise, we will explore dynamic structural movements, which could provide additional aerodynamic benefits by eliminating vortex-induced vibrations \citep{Galvao08}. 
While the present study was confined to a laboratory setting, work is underway to demonstrate full-size morphing vehicles in the field. 
Future studies should also investigate the potential benefits of morphing for various vehicle models, such as commercial trucks, which could bring even greater environmental and economic benefits \citep{McTavish21}.

%%%%%%%%%%%%%%%%%%%%%%%%%%%%%%%%%%%%%%%%%%%%%%%%%%%%%%%%%%%%%%%%%%%%%%
%%%%%%%%%%%%%%% begin table   %%%%%%%%%%%%%%%%%%%%%%%%%%
%\begin{table}[t]
%\caption{Figure and table captions do not end with a period}
%\begin{center}
%\label{table_ASME}
%\begin{tabular}{c l l}
%& & \\ % put some space after the caption
%\hline
%Example & Time & Cost \\
%\hline
%1 & 12.5 & \$1,000 \\
%2 & 24 & \$2,000 \\
%\hline
%\end{tabular}
%\end{center}
%\end{table}
%%%%%%%%%%%%%%%% end table %%%%%%%%%%%%%%%%%%% 
%%%%%%%%%%%%%%%%%%%%%%%%%%%%%%%%%%%%%%%%%%%%%%%%%%%%%%%%%%%%%%%%%%%%%%

%%%%%%%%%%%%%%%%%%%%%%%%%%%%%%%%%%%%%%%%%%%%%%%%%%%%%%%%%%%%%%%%%%%%%%
\begin{acknowledgment}
The authors gratefully acknowledge Dr.~Ahmad Vasel-Be-Hagh for designing, constructing, and providing detailed operational instructions for the wind tunnel facility at Tennessee Technological University, which made this study possible.
\end{acknowledgment}

%%%%%%%%%%%%%%%%%%%%%%%%%%%%%%%%%%%%%%%%%%%%%%%%%%%%%%%%%%%%%%%%%%%%%%

\begin{nomenclature}
\entry{A}{Frontal area of the vehicle model, $\mathrm{m^2}$}
\entry{$C_{\mathrm{D}}$}{Drag coefficient; $\frac{ 2 \overline{F_{\mathrm{D}}} }{   \rho_{\mathrm{a}} U^2 A}$}
\entry{$E$}{Young's modulus, $\mathrm{Pa}$}
\entry{$F_D$}{Drag force, $\mathrm{N}$}
\entry{$\overline{F_{\mathrm{D}}}$}{Mean drag force over the acquisition period, $\mathrm{N}$}
\entry{$g$}{Gravitational acceleration, $\mathrm{m/s^2}$}
\entry{$h$}{Half-thickness of the morphing panels, $\mathrm{m}$}
\entry{$H$}{Height of the vehicle, $\mathrm{m}$}
\entry{$l$}{Total length of morphing panels, $\mathrm{m}$}
\entry{$L$}{Length of the vehicle, $\mathrm{m}$}
\entry{$\Rey$}{Reynolds number; $\Rey = L U / \nu$}
\entry{$U$}{Upstream wind speed, $\mathrm{m/s}$}
\entry{$w$}{Width of morphing panels, $\mathrm{m}$}
\entry{$\gamma$}{Permissible relative deflection}
\entry{$\delta_{\mathrm{max}}$}{Maximum panel deflection, $\mathrm{m}$}
\entry{$\theta_1$, $\theta_2$, and $\theta_3$}{Local turning angles of the panels, rad}
\entry{$\Theta$}{Domain of feasible design variables}
\entry{$\nu$}{Kinematic viscosity of air, $\mathrm{m^2/s}$}
\entry{$\rho$}{Density of the morphing panels, $\mathrm{kg/m^3}$}
\entry{$\rho_{\mathrm{a}}$}{Density of air, $\mathrm{kg/m^3}$}
\entry{$\sigma_{\mathrm{max}}$}{Maximum stress across panel thickness, $\mathrm{Pa}$}
\entry{$\sigma_{\mathrm{u}}$}{Ultimate tensile strength, $\mathrm{Pa}$}
\end{nomenclature}

%%%%%%%%%%%%%%%%%%%%%%%%%%%%%%%%%%%%%%%%%%%%%%%%%%%%%%%%%%%%%%%%%%%%%%

\bibliographystyle{asmems4}

% Here's where you specify the bibliography database file.
% The full file name of the bibliography database for this
% article is asme2e.bib. The name for your database is up
% to you.
\bibliography{citation_morphing}

%%%%%%%%%%%%%%%%%%%%%%%%%%%%%%%%%%%%%%%%%%%%%%%%%%%%%%%%%%%%%%%%%%%%%%
%\appendix       %%% starting appendix
%\section*{Appendix A: Head of First Appendix}
%Avoid Appendices if possible.
%
%%%%%%%%%%%%%%%%%%%%%%%%%%%%%%%%%%%%%%%%%%%%%%%%%%%%%%%%%%%%%%%%%%%%%%%
%\section*{Appendix B: Head of Second Appendix}
%\subsection*{Subsection head in appendix}
%The equation counter is not reset in an appendix and the numbers will
%follow one continual sequence from the beginning of the article to the very end as shown in the following example.
%\begin{equation}
%a = b + c.
%\end{equation}

%%%%%%%%%%%%%%%%%%%%%%%%%%%%%%%%%%%%%%%%%%%%%%%%%%%%%%%%%%%%%%%%%%%%%%
\end{document}